\documentclass{article}

\usepackage[preprint]{neurips_2024}

\usepackage[utf8]{inputenc} 
\usepackage[T1]{fontenc}    
\usepackage{hyperref}       
\usepackage{url}            
\usepackage{booktabs}       
\usepackage{amsfonts}       
\usepackage{nicefrac}       
\usepackage{microtype}      
\usepackage{xcolor}         

\usepackage{multirow}
\usepackage{amsmath}
\usepackage{multicol}
\usepackage{graphicx}
\usepackage{enumitem}
\usepackage{bbm}
\usepackage{pifont}
\usepackage{amsmath}
\usepackage{subcaption}
\usepackage{caption}
\usepackage{tabularx}

\usepackage{graphicx}
\usepackage{graphics}   
\usepackage{amssymb}
\usepackage{balance}
\usepackage{adjustbox}
\usepackage{algorithm}
\usepackage{algorithmic}

\usepackage{color, colortbl}

\definecolor{Gray}{gray}{0.9}
\newcommand{\ceil}[1]{\left\lceil #1 \right\rceil}
\newcommand{\eg}{\textit{e}.\textit{g}.}
\newcommand{\ie}{\textit{i}.\textit{e}.}
\newcommand{\etal}{et al.}

\newcommand{\bcmark}{\textcolor{black}{\ding{51}}} 
\newcommand{\bxmark}{\textcolor{black}{\ding{55}}} 

\definecolor{lblue}{rgb}{0, 0.2, 0.8}
\definecolor{orange}{rgb}{1.0, 0.5, 0.0}

\definecolor{lblue}{rgb}{0, 0.2, 0.8}
\definecolor{dorange}{rgb}{0.8, 0.4, 0.0}

\definecolor{LightCyan}{rgb}{0.88,1,1}
\definecolor{Custompink}{rgb}{1,0.93,0.93}
\newcolumntype{g}{>{\columncolor{Custompink}}c}

\title{GroupCoOp: Group-robust Fine-tuning via\\Group Prompt Learning}

\author{
Nayeong Kim$^{1}$ \quad
Seong Joon Oh$^{2}$ \quad
Suha Kwak$^{1}$ \\
$^{1}$ Pohang University of Science and Technology (POSTECH), South Korea\\
$^{2}$Tübingen AI Center, Universität Tübingen\\
}

\begin{document}

\maketitle

\begin{abstract}
Parameter-efficient fine-tuning (PEFT) of vision-language models (VLMs) excels in various vision tasks thanks to the rich knowledge and generalization ability of VLMs.
However, recent studies revealed that such fine-tuned VLMs are vulnerable to spurious correlations stemming from the subgroup imbalance in the fine-tuning datasets.
To resolve this issue, we propose \textbf{Group} \textbf{Co}ntext \textbf{Op}timization (\textbf{GroupCoOp}), a simple and effective debiased fine-tuning algorithm that enhances the group robustness of fine-tuned VLMs.
Its key idea is to employ group-specific text prompts as group representatives serving as multiple classifiers for their target class.
The rich semantic knowledge of the text encoder of VLM enables the discovery of effective group prompts even for groups with a small number of training samples.
Leveraging the group prompts for each class addresses the issues caused by the group-imbalanced training set, such as the neglect of minority groups and the scattered distribution of each class in the embedding space.
GroupCoOp achieved the best results on five benchmarks across five CLIP architectures and occasionally outperformed prior methods that fine-tune the entire network, despite training only 0.016\% of the network's parameters.
\end{abstract}
\section{Introduction}
Vision-language models (VLMs) have been known for their rich knowledge and generalization capability, acquired by leveraging textual descriptions alongside images for training~\citep{brown2020language, jia2021scaling, li2022blip, radford2021learning, singh2022flava,li2022supervision}. 
These features enable VLMs to demonstrate outstanding performance across various downstream vision tasks, surpassing models pretrained solely on images~\citep{tan2019efficientnet,xie2020self, chen2020simple,mahajan2018exploring, touvron2019fixing, kolesnikov2019large, dosovitskiy2021image, chen2020big, BYOL, resnet}. 
In this context, parameter-efficient fine-tuning (PEFT) has emerged as a promising way of adapting VLMs to diverse vision tasks while keeping their rich knowledge and generalization ability~\citep{kumar2022fine, radford2021learning, gao2024clip, zhou2022learning, zhang2022tip}.

One may expect PEFT to ensure \emph{group robustness} of fine-tuned VLMs, \ie, their robustness to spurious correlations between true labels and irrelevant attributes arising from subgroup imbalance of training data~\citep{geirhos2020shortcut,sagawa2019distributionally}, thanks to their rich knowledge gained from a vast amount of data during pretraining.
However, empirical observations suggest that fine-tuned VLMs are vulnerable to  
such spurious correlations~\citep{zhang2022contrastive}.
For example, VLMs fine-tuned for bird species classification proficiently classify waterbird images against a water background (majority group) but often fail to handle images of the same class but with a land background (minority group), as they are prone to taking undesirable shortcuts stemming from spurious correlations between bird classes and background during fine-tuning.
Moreover, fine-tuning VLMs through empirical risk minimization (ERM) has been known to exacerbate the bias~\citep{zhang2022contrastive}.

Debiased training algorithms~\citep{sagawa2020investigation, sagawa2019distributionally, JTT, LfF, bahng2020learning, kim2022learning, kirichenko2022last, nam2022spread, zhang2022correct, wu2023discover} have been studied to improve group robustness of deep neural networks trained on biased datasets. 
A common approach in this line of work is group-wise reweighting and group-wise resampling to ensure balanced learning among groups~\citep{sagawa2020investigation, sagawa2019distributionally, JTT, LfF, kim2022learning, kirichenko2022last}.
However, most of these methods were originally designed with the assumption of training the entire model, and thus, they are prohibitively expensive to apply to VLMs. 
There are only a few debiasing methods that tune only a small portion of parameters, \eg, the classifier on top of a frozen backbone network~\citep{kirichenko2022last, kim2022learning}, but they do not perform well with VLMs as demonstrated in Table~\ref{tab:performance} since they assume the backbone network tuned for the target dataset in advance, which is infeasible for VLMs.

Recently, several studies have addressed the group robustness of VLMs through PEFT~\citep{zhang2022contrastive, yang2023mitigating, you2024calibrating, dehdashtian2024fairerclip, zhang2024amend}. 
They attribute the problem to scattered distributions of images from the same class but different groups in the embedding space.
To address this, they transform visual features of images of the same class so that they are aligned with predefined class-specific textual features by an adapter~\citep{zhang2022contrastive} or projection layers~\citep{yang2023mitigating, you2024calibrating, zhang2024amend}. Additionally, some studies employ kernel methods to transform both text and visual features jointly~\citep{dehdashtian2024fairerclip}.
However, these methods often perform worse than ERM-based PEFT, since training such feature transformations is challenging for minority groups that lack training samples.

Instead of transforming visual features, we propose \textit{Group Context Optimization} (GroupCoOp), a simple and effective debiased PEFT algorithm.
GroupCoOp aims to learn a group representative for each group in the embedding space \emph{without group annotation}, and the group representative serves as a classifier for the associated group.
GroupCoOp is inspired by prompt tuning, which provides task context through text input along with class names, leading to state-of-the-art performance across various downstream tasks without fine-tuning an embedding space of VLM~\citep{radford2021learning, zhou2022learning}.
Similarly, GroupCoOp learns group-specific text prompts and conducts group classification of input by identifying the group with the text prompt closest to the input in the embedding space of the VLM.
Then the class label of the identified group prompt in the embedding space is used as the final classification result.
GroupCoOp ensures balanced learning among groups by learning group prompts for all groups, including minority groups, thus addressing the common issue of minority neglect when training on group-imbalanced datasets.
By utilizing multiple group representatives per class, GroupCoOp effectively manages the scattered distribution of each class. 
This strategy enables the model to capture the diverse characteristics within each class, leading to more robust and consistent performance across all subgroups.

In addition, we propose a pseudo group labeling strategy with frozen VLMs to address the challenge of obtaining group labels for training sets. 
Following Nam~\etal~\cite{nam2022spread}, we leverage the group-labeled validation set not only for hyperparameter tuning, which is common in this research field, but also for assigning pseudo group labels to the training set. 
We employ confidence-based pseudo labeling~\cite{jung2022learning} to assign training data spurious attribute labels, which are then combined with class labels to generate pseudo group labels of the data. 

We extensively evaluated our method on four benchmarks with five architectures, where it outperformed every prior art. The main contribution of this paper is threefold:
\vspace{-2mm}
\begin{itemize}[leftmargin=6mm] 
    \setlength\itemsep{-0.2mm}
    \item We present GroupCoOp, a new debiased fine-tuning algorithm for frozen VLMs with no group labels. Context optimization allows GroupCoOp to effectively improve group robustness while preserving a rich representation of VLMs.
    \item We propose generating pseudo group labels of the training set with frozen VLMs using a labeled validation set, a method not previously explored in debiased PEFT.
    \item GroupCoOp achieved state-of-the-art results on five datasets using five architectures without training data group labels.
    Furthermore, GroupCoOp demonstrated higher performance with only 0.016\% learnable parameters compared with debiasing methods based on full fine-tuning.
\end{itemize}

\begin{figure*}[t!]
\centering
\includegraphics[width=\linewidth]{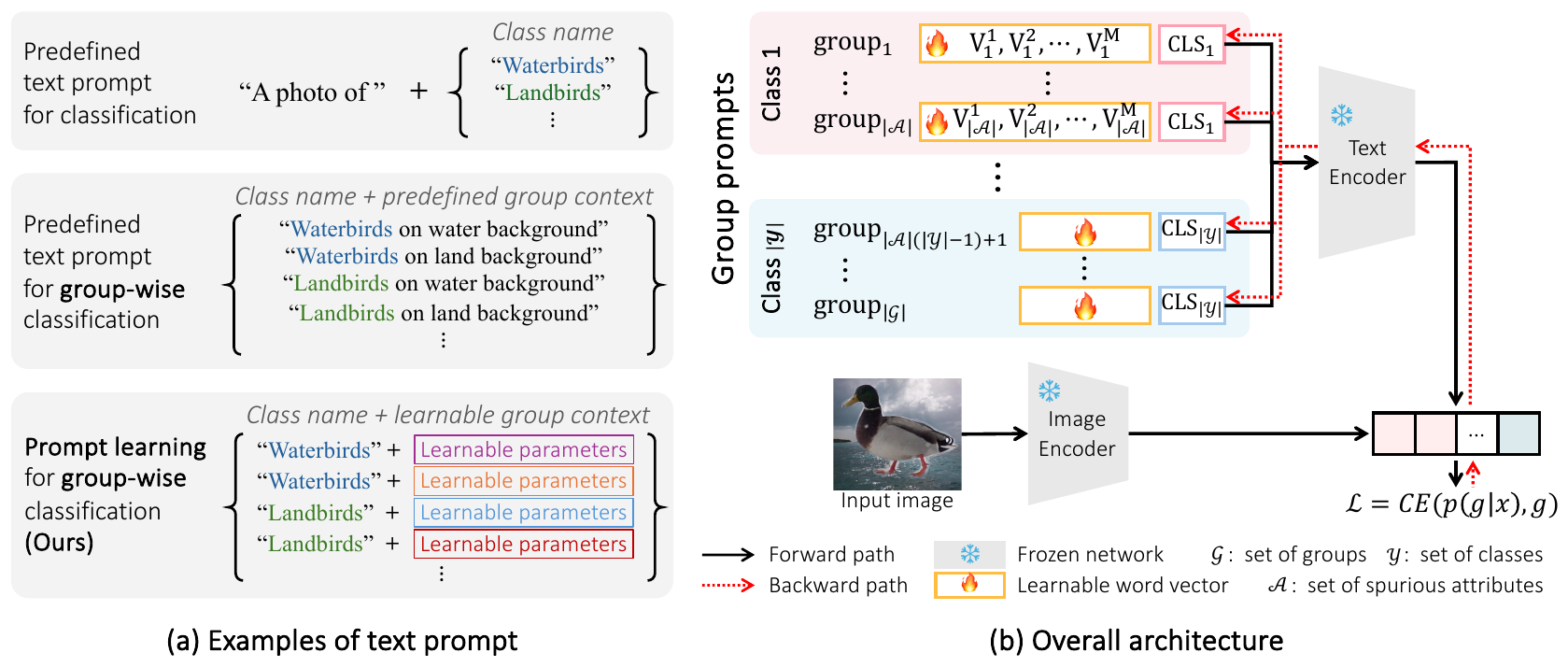}
\vspace{-5mm}
\caption{
An illustration of group context learning (a) and the overall framework of GroupCoOp (b). 
(a) compares different text prompt strategies: predefined prompts for classification, predefined prompts for group-wise classification, and prompt learning for group-wise classification (ours). While predefined prompts rely on fixed contexts, prompt learning replaces them with adaptive, learnable parameters. 
(b) presents the fine-tuning phase involving learning group contexts, where group-specific prompts are fed into the frozen text encoder, alongside image features extracted by the frozen image encoder. The loss is calculated as cross-entropy between the predicted and actual (pseudo) group labels, with backpropagation occurring through the text encoder.  
}
\label{fig:architecture}
\end{figure*}

\section{Related Work}
\subsection{Debiased training}
A body of research has been proposed to improve group robustness of a model by mitigating bias stemming from spurious correlation in biased training set~\citep{arjovsky2019invariant,bahng2020learning,sagawa2019distributionally, sagawa2020investigation,yao2022improving, tartaglione2021end, hwang2022selecmix, kirichenko2022last, kim2023removing}. 
To balance between groups, reweighting~\citep{sagawa2019distributionally, kim2023removing}, resampling~\citep{sagawa2020investigation, kirichenko2022last}, or mixup~\citep{yao2022improving} between groups have been introduced.
However, these methods require group label which is impractical because such supervision is costly.
To overcome this challenge, debiased training algorithms without group labels have been proposed~\citep{LfF, JTT, lee2021learning, nam2022spread, jung2022learning, kim2022learning, zhang2022correct, hwang2022selecmix, wu2023discover}. 
A common approach to replace ground truth group label for debiased training is leveraging the correctness of model prediction and class label as a criterion for grouping~\citep{LfF, JTT, lee2021learning, kim2022learning, zhang2022correct, hwang2022selecmix}. 
SSA~\citep{nam2022spread} and CGL~\citep{jung2022learning} proposed to train a group classifier using a partially group annotated set for pseudo group labeling. DFR~\citep{kirichenko2022last} retrain the last layer using labeled validation set atop ERM pretrained representation.
In other hand, DISC~\citep{wu2023discover} and yang~\etal~\citep{yang2023mitigating} introduced bias discovery algorithm by utilizing extra knowledge~\citep{wu2023discover, yang2023mitigating}.
However, these methods were designed with the assumption of training the entire model.

Only a few recent studies~\citep{zhang2022contrastive, yang2023mitigating, dehdashtian2024fairerclip, you2024calibrating, zhang2024amend} addressed improving group robustness for frozen VLMs. 
Zhang and R{\'e}~\cite{zhang2022contrastive} introduced an adapter atop the image encoder and contrastive adapting to learn samples belonging to the same class but different groups to be closer in the embedding space. 
Yang~\etal~\cite{yang2023mitigating} proposed to
fine-tune only the projection layer of VLMs by contrastive learning. 
CFR~\citep{you2024calibrating} retrained the projection layer and utilized ERM-tuned CLIP to replace group annotations.
FairerCLIP~\citep{dehdashtian2024fairerclip} introduced jointly debiasing image and text features in reproducing kernel Hilbert spaces (RKHSs).
CoOPood~\citep{zhang2024amend} trained a learnable prompt and projection layer to pull the text embedding closer to the invariant image embedding and push the text embedding away from the spurious image embedding.
However, their method showed worse performance compared with SOTA full fine-tuning debiasing methods.
In contrast, we propose learning a group representative vector within the VLMs' embedding space rather than constructing a new embedding space for the downstream task. Our method outperformed both existing debiasing methods for VLMs and full fine-tuning debiasing methods. 

\subsection{Parameter-efficient transfer learning}
Parameter-efficient transfer learning~\citep{chen2022adaptformer, he2021towards, houlsby2019parameter, hu2021lora, jia2022visual, lian2022scaling, pfeiffer2020adapterfusion,rebuffi2017learning} has emerged as a crucial technique for fine-tuning pretrained large models by minimizing computational resources and preventing catastrophic forgetting, focusing on modifying only a small subset of parameters rather than updating a large portion of the model.
One popular approach involves the use of adapter modules~\citep{chen2022adaptformer, houlsby2019parameter, pfeiffer2020adapterfusion,luo2023towards}, which are lightweight, trainable layers inserted into pretrained networks, allowing for efficient adaptation without altering the entire mode.
Another technique is prompt-based learning~\citep{lester2021power, li2021prefix, wang2022learning, smith2023coda}, where a limited number of tunable prompts guide the model during fine-tuning, leveraging the pretrained knowledge within the model to enable efficient adaptation with minimal parameter updates.
We propose a group-robust parameter-efficient fine-tuning method that leverages the extensive language knowledge encoded in VLMs to address the group imbalance.

\section{Problem Setting}
We consider a classification task that, given an image $x\in\mathcal{X}$ as input, predicts its class label $y\in\mathcal{Y}$ by leveraging the similarity between text prompts using class name $\texttt{CLS}$ and the input $x$ in the embedding space of a VLM~\citep{zhang2022contrastive, yang2023mitigating}.
Specifically, our focus lies in scenarios with imbalanced subgroups that result in a spurious correlation between class labels and subgroups $g\in\mathcal{G}$.
Typically, these groups are defined by a combination of the class label $y\in\mathcal{Y}$ and a spurious attribute $a\in\mathcal{A}$, denoted as $\mathcal{G}=\mathcal{Y}\times\mathcal{A}$.
For each class, larger groups are referred to as \textit{majority groups} and the smaller groups are referred to as \textit{minority groups}.
Since ERM tends to exploit undesirable shortcuts from spurious correlations arising from the subgroup imbalance, a model prone to classify well on samples from the majority group but not well on the minority group.
Suppose the dataset consists of $N$ samples, each of which comprises inputs $x\in\mathcal{X}$, class labels $y\in\mathcal{Y}$, and group labels $g\in\mathcal{G}$; the group label is used for evaluation and is not provided during training.
$P_g$ is the distribution conditioned on $g$ for any $g\in\mathcal{G}$. 
To evaluate the group robustness, we calculate the worst-case group accuracy following prior work~\citep{sagawa2019distributionally,zhang2022contrastive}:
\begin{align}
    \mathcal{E}_{\textsc{Worst}} := \min_{g \in \mathcal{G}} \mathbb{E}_{(x, y, g) \sim P_g}[\mathbb{I}(\hat{y} = y)],
\end{align}
where $\mathbb{I}$ denotes indicator function and $\hat{y}$ denotes class prediction.
In this paper, our goal is to improve group robustness of VLMs fine-tuned on such imbalanced subgroups and evaluate it using the worst group accuracy.

\section{Proposed method}
We propose GroupCoOp, a simple and effective debiasing algorithm for PEFT of VLMs based on prompt learning~\citep{zhou2022learning}. 
GroupCoOp aims to fine-tune frozen VLMs to perform well across all groups with a tiny number of additional learnable parameters while preserving the rich representation of the VLMs.
The overall process of GroupCoOp is illustrated conceptually in Figure~\ref{fig:architecture}.
The rest of this section first introduces the group context optimization (Section~\ref{sec:group_context_optimization}), and then class inference with group prompts (Section~\ref{sec:inference}, and the pseudo group labeling 
(Section~\ref{sec:pseudo_labeling}).

\subsection{Group context optimization}\label{sec:group_context_optimization}
Assuming group labels are available during training, we fine-tune frozen VLMs to perform well across all groups by learning the context for each group.
To this end, we introduce a text prompt $t_i$ for group $i$ using $M$ learnable group context word vectors $\{V_i^1,\dots,V_i^M\}$ and the word vector of the corresponding class name $\texttt{CLS}_c$:
\begin{align}
    t_i=[V_i^1, V_i^2, \cdots, V_i^M, \texttt{CLS}_c], \quad  \forall i\in\{1, \dots,|\mathcal{G}|\}
\end{align}
where each $V_i^m\in \mathbb{R}^D$ is of the same dimension as the word embedding, $M$ denotes the number of learnable tokens, and $c$ equals $\ceil{i/|\mathcal{A}|}$. 
Then the group-wise prompt $t$ is forwarded to the text encoder $f(\cdot)$, where the embedding of prompt $f(t)$ serves as the group classifier.  
Subsequently, the group prediction probability is calculated from the cosine similarity between the group prompt and the input image in the embedding space:
\begin{align}
    p(g_i|x)=\frac{\exp(\cos(f(t_i), h(x))}{\sum_{k=1}^{|\mathcal{G}|}\exp(\cos(f(t_k), h(x))},
\end{align}
where $f(\cdot)$ denotes the text encoder and $h(\cdot)$ denotes the image encoder.
Then, group prompts are trained by minimizing the cross-entropy loss for group classification:
\begin{align}
    \mathcal{L}_{\texttt{CE}}= - \frac{1}{|\mathcal{B}|}\sum_{(x,y,g)\in \mathcal{B}} g\log(p(g|x)),
\end{align}
where $\mathcal{B}$ denotes a mini-batch that is balanced to groups. 
Since samples from the same group share class features and spurious features, they are close to each other in the embedding space. 
The context word vectors $V_i$ learn group contexts by minimizing the group cross-entropy loss with error backpropagation through the text encoder.
This allows GroupCoOp to leverage the language knowledge embedded in the text encoder effectively. 
In particular, when the groups are significantly unbalanced in size, and thus minor groups have extremely few samples, leveraging the knowledge of the text encoder is highly beneficial for learning the group context for the minority groups. This allows GroupCoOp to perform well even in extreme scenarios where a minority group has only a single sample.

\subsection{Class inference with group prompts}\label{sec:inference}
During inference, the model makes class predictions based on group predictions, as each group is a member of a class. To this end, we first predict the group ID of the input and then consider the class of the predicted group as the classification result. 
If the predicted group ID is $i$, then the class of the predicted group is given by $\ceil{i/|\mathcal{A}|}$.

\subsection{Pseudo group labeling in VLMs}\label{sec:pseudo_labeling}
Before training, we generate pseudo group labels for training data. 
Following Nam~\etal~\cite{nam2022spread}, the group-labeled validation set is utilized not only for hyperparameter tuning, which is common in this research field, but also for assigning pseudo group labels to the training set. 
To assign pseudo group labels, we first utilize the confidence-based pseudo labeling~\citep{jung2022learning} with the group-labeled validation set to generate pseudo spurious attribute labels $a_{\text{pseudo}}\in\mathcal{A}$. Subsequently, pseudo group labels $g_{\text{pseudo}}\in \mathcal{G}$ are assigned using these pseudo spurious attribute labels in conjunction with the class labels $y$.

This process begins by splitting the group-labeled validation set $D^{\text{val}}$ into $D_{\text{s}}^{\text{val}}$ for training a spurious attribute classifier $s(\cdot)$ and $D_{\tau}^{\text{val}}$ for determining threshold $\tau$.
The spurious attribute classifier is trained using the image feature $h(x)$ of each sample $x$ from  $D_{\text{s}}^{\text{val}}$ as input. 
To reduce the effects of incorrect pseudo labels, the prediction of $s(\cdot)$ is used only for samples whose confidence scores exceed a threshold $\tau$.
$\tau$ is determined to optimize the accuracy in predicting spurious attributes of $D_{\tau}^{\text{val}}$ by $s(\cdot)$. 
This approach is inspired by a threshold-based strategy in the out-of-distribution sample detection task~\citep{hendrycks2017baseline}.
The optimization problem is defined as follows:
\begin{align}
    \max_\tau 
    \biggl(
    \sum_{\substack{x\in\\\{\max s(h(x)) >\tau\}}}\mathbb{I}\left(\arg\max s(h(x)) = a\right) \nonumber 
    +\sum_{\substack{x\in\\\{\max s(h(x)) \le\tau\}}}\mathbb{I}(\arg\max s(h(x)) \neq a )
    \biggr)
\end{align}
Here, $\tau=1$ corresponds to assigning a `random label' to all samples, while $\tau=0$ aligns with the vanilla pseudo-labeling strategy.

Then, samples whose confidence scores exceed $\tau$ are assigned pseudo labels predicted by this classifier, and the others are labeled at random:
\begin{align}
    a_{\text{pseudo}} = 
    \begin{cases}
        \arg\max s(h(x)) & \text{if } \max s(h(x)) > \tau, \\
        \text{random sampling} & \text{otherwise}.
    \end{cases}
\end{align}
In contrast to the method of Jung~\etal~\cite{jung2022learning}, which employs the empirical distribution $P(A|Y=y)$ for sampling, we adopt random sampling since the class conditional distribution estimated from the validation set may not well approximate that of the training set.

The pseudo group label $g_\text{pseudo}\in\{1,...,|\mathcal{G}|\}$ is determined by the pseudo spurious attribute label $a_{\text{pseudo}}\in\{1,...,|\mathcal{A}|\}$ and the class label $y$ as follows:
\begin{align}
     g_{\text{pseudo}} = |\mathcal{A}|\cdot(y-1)+a_{\text{pseudo}},
\end{align}
where $y\in\{1,...,|\mathcal{Y}|\}$ denotes the ground truth class label of the sample $x$. 
We empirically demonstrated the effectiveness of the proposed pseudo group labeling by comparing it with other pseudo group labeling methods; details are presented in Section~\ref{sec:effectiveness_of_pseudo}.

\section{Experiments}
\subsection{Setup}
\noindent\textbf{Implementation details.} 
We adopt five pretrained CLIP models \{RN50, RN101, ViT-L/14, ViT-B/16, ViT-B/32\}~\citep{radford2021learning} using publicly available weights as a frozen backbone. 
For computational efficiency, we save image features once before training and use these features as visual inputs during training.
We set the learning rate to \{0.0002, 0.001, 0.5, 0.2, 0.002\} and the optimizer \{Adam, Adam, Adam, SGD, Adam\}, respectively for \{Waterbirds, CelebA, Metashift, BREEDS Living17, BREEDS Nonliving26\}. 
The batch size is set to 256, and the number of learnable word vectors $M$ is set to 4. We run GroupCoOp on 3 random seeds and report the average and the standard deviation.
For generating pseudo group labels, the spurious attribute classifier $s(\cdot)$ is implemented as a single linear layer.
All experiments are conducted on a single NVIDIA RTX 3090 GPU.

\noindent\textbf{Evaluation metrics.} The quality of group robustness of debiased training algorithm is measured by \textsc{Worst}, which denotes the minimum group average accuracy. Following Sagawa~\etal~\cite{sagawa2019distributionally}, we also include \textsc{InDist}, which represents the weighted average of group accuracy, where the weights correspond to the relative proportions of group sizes in the training data. 
A high \textsc{Worst} accuracy indicates that a model is unbiased to groups, while a biased model attains a low \textsc{Worst} but may have high \textsc{InDist}.

\subsection{Datasets}
\noindent\textbf{Waterbirds}~\citep{CUB200, sagawa2019distributionally}. Waterbirds benchmarks spurious correlation between object and background. We use bird species ($\mathcal{Y}=\{\text{waterbird}, \text{landbird}\}$) as target class and background ($\mathcal{A}=\{\text{water background},\text{land background}\}$) as the spurious attribute.
In this dataset, the CLIP zero-shot results are highly biased, and since the training set is also biased, ERM fine-tuning is likely to result in even more bias.

\noindent\textbf{CelebA}~\citep{CelebA, sagawa2019distributionally}. Following the previous settings~\citep{sagawa2019distributionally}, we consider spurious correlation between the label and demographic information. We use hair color ($\mathcal{Y}=\{\text{blond}, \text{black}\}$) as target class and gender ($\mathcal{A}=\{\text{male},\text{female}\}$) as the spurious attribute. 
In this dataset, CLIP zero-shot results are less biased, but the overall performance on the dataset is low, necessitating fine-tuning. However, due to the bias in the training set, ERM fine-tuning can easily result in greater bias compared to zero-shot.

\noindent\textbf{Metashift}~\citep{liang2022metashift}. Metashift benchmarks spurious correlation between object and background in natural images. Following the previous setting~\citep{zhang2024amend}, we use $\mathcal{Y}=\{\text{Cat}, \text{Dog}\}$ as target class and $\mathcal{A}=\{\text{indoors}, \text{outdoors}\}$ as spurious attributes.

\noindent\textbf{BREEDS} (Living17, Nonliving26)~\citep{santurkar2021breeds, zhang2022contrastive}. BREEDS Living17 and BREEDS Nonliving26 benchmark sub-population shifts between \texttt{source} dataset and \texttt{target} dataset. While the original benchmarks evaluate a model on unseen \texttt{target} groups, we adopt the group robustness setting of BREEDS as proposed by Zhang and R{\'e}~\cite{zhang2022contrastive}. 
Both the training set and the test set contain both \texttt{source} groups and \texttt{target} groups. 
BREEDS Living17 and BREEDS Nonliving26 contain 17 and 26 classes, respectively, with 4 subgroups for each class.
In this dataset, the CLIP zero-shot results are highly biased, but since the training set is group-balanced, ERM fine-tuning can alleviate some of this bias. The experimental results on BREEDS validate whether the debiasing algorithm can be effectively applied not only in the presence of dataset bias but also in its absence.

\subsection{Baselines} 
We compare GroupCoOp with zero-shot inference methods: Zero-shot inference with text prompt (class-wise and group-wise), and KNN using image features.
We also compare GroupCoOp with PEFT baselines: ERM Linear, ERM Adapter, and WiSE-FT.
CoOp~\citep{zhou2022learning}, which trains a unified learnable prompt, and CoOp-CSC, which learns class-specific prompts.
Subsequently, the previous debiasing algorithms (LWBC~\citep{kim2022learning},  DFR~\citep{kirichenko2022last}, GroupDRO~\citep{sagawa2019distributionally}, JTT~\citep{JTT}, and CnC~\citep{zhang2022correct}) were applied to the frozen VLMs, training only the linear layer after the frozen image encoder without any backbone training step.
Lastly, we compared GroupCoOp with debiased fine-tuning algorithms for VLMs: 
Orth-Cali~\citep{chuang2023debiasing} projects text embeddings using only class and group names to yield debiased text embeddings without images.
CA~\citep{zhang2022contrastive} trains an adapter with group-wise contrastive learning. 
Yang~\etal~\cite{yang2023mitigating} and CFR~\citep {you2024calibrating} retrain projection layers of the image encoder through group-wise contrastive learning and a curated calibration set, respectively.
FairerCLIP~\citep{dehdashtian2024fairerclip} employs kernel methods on top of image and text features.
CoOPood~\citep{zhang2024amend} trains a unified learnable prompt and projection layers of image and text features.

\begin{table*}[th!]
  \caption{\textsc{Worst} and \textsc{InDist} metrics (\%) evaluated on the Waterbirds and CelebA datasets. We mark the best and the second-best \textsc{Worst} in \textbf{bold} and \underline{underline}, respectively.}
  \label{tab:performance}
  \centering
  \small
  \resizebox{\linewidth}{!}{
  \begin{tabular}{lcgcgcgcgc}
  \toprule
  & & \multicolumn{4}{c}{CLIP RN50} & \multicolumn{4}{c}{CLIP ViT-L/14}\\
   & Bias & \multicolumn{2}{c}{Waterbirds} & \multicolumn{2}{c}{CelebA} & \multicolumn{2}{c}{Waterbirds} & \multicolumn{2}{c}{CelebA} \\
    Method & label & \textsc{Worst} & \textsc{InDist} & \textsc{Worst} & \textsc{InDist} & \textsc{Worst} & \textsc{InDist} & \textsc{Worst} & \textsc{InDist} \\
    \midrule
    Zero-shot (class) & \bxmark & 36.6$_{\pm {0.0}}$ & 92.2$_{\pm {0.0}}$ & 74.0$_{\pm {0.0}}$ & 81.9$_{\pm {0.0}}$ & 25.7$_{\pm {0.0}}$ & 87.3$_{\pm {0.0}}$ & 62.1$_{\pm {0.0}}$ & 71.9$_{\pm {0.0}}$\\
    
    Zero-shot (group) & \bxmark & 55.9$_{\pm {0.0}}$ & 87.8$_{\pm {0.0}}$ & 70.8$_{\pm {0.0}}$ & 82.6$_{\pm {0.0}}$ & 27.4$_{\pm {0.0}}$ & 85.5$_{\pm {0.0}}$ & 72.4$_{\pm {0.0}}$ & 81.8$_{\pm {0.0}}$\\

    KNN & \bxmark & 43.0$_{\pm {0.0}}$ & 96.1$_{\pm {0.0}}$ & 24.4$_{\pm {0.0}}$ & 91.0$_{\pm {0.0}}$ & 63.6$_{\pm {0.0}}$ & 97.1$_{\pm {0.0}}$ & 18.9$_{\pm {0.0}}$ & 89.6$_{\pm {0.0}}$\\ 
    
    ERM Linear  & \bxmark & 61.2$_{\pm {1.5}}$ & 96.4$_{\pm {0.0}}$ & 44.6$_{\pm {5.0}}$ & 95.4$_{\pm {0.0}}$ & 65.4{$_{\pm {0.5}}$} & 97.7{$_{\pm {0.1}}$} & 30.4{$_{\pm {1.5}}$} & 94.6{$_{\pm {0.1}}$} \\ 
    
    ERM Adapter & \bxmark & 63.0$_{\pm {0.4}}$ & 96.0$_{\pm {1.1}}$ & 41.9$_{\pm {4.5}}$ & 94.7$_{\pm {0.4}}$ & 76.1{$_{\pm {1.8}}$} & 97.8{$_{\pm {0.1}}$} & 40.0{$_{\pm {5.6}}$} & 94.3{$_{\pm {0.3}}$} \\ 
    
    WiSE-FT & \bxmark & 49.8$_{\pm {0.0}}$ & 91.0$_{\pm {0.0}}$ & 85.6$_{\pm {0.0}}$ & 88.6$_{\pm {0.0}}$ & 65.9$_{\pm {0.0}}$ & 97.6$_{\pm {0.0}}$ & 80.0$_{\pm {0.0}}$ & 87.4$_{\pm {0.0}}$\\

    CoOp & \bxmark & 63.4$_{\pm {0.9}}$ & 96.3$_{\pm {0.1}}$ & 44.3$_{\pm {2.9}}$ & 95.2$_{\pm {0.1}}$ & 76.8$_{\pm {1.0}}$ & 97.2$_{\pm {0.2}}$ & 36.1$_{\pm {2.8}}$ & 94.3$_{\pm {0.2}}$\\
    
    CoOp-CSC & \bxmark & 65.0$_{\pm {2.3}}$ & 96.1$_{\pm {0.1}}$ & 43.5$_{\pm {3.9}}$ & 95.2$_{\pm {0.1}}$ & 74.4$_{\pm {2.0}}$ & 97.2$_{\pm {0.2}}$ & 34.8$_{\pm {2.6}}$ & 94.5$_{\pm {0.1}}$\\
    
    GroupDRO & \bcmark & 75.1\phantom{$_{\pm {0.0}}$} & 83.8\phantom{$_{\pm {0.0}}$} & 84.1\phantom{$_{\pm {0.0}}$} & 89.5\phantom{$_{\pm {0.0}}$} & - & - & - & -\\  
    LWBC & \bxmark &67.8$_{\pm 2.0}$ & 96.2$_{\pm 0.2}$ & 77.8\phantom{$_{\pm {0.0}}$} & 87.4\phantom{$_{\pm {0.0}}$} & 78.0$_{\pm {5.1}}$ & 96.9$_{\pm {0.6}}$ & 72.6$_{\pm {1.8}}$ & 86.6$_{\pm {3.6}}$\\
    
    DFR (Sub) & \bxmark & 66.1$_{\pm {5.5}}$ & 92.9$_{\pm {2.2}}$ & 80.9$_{\pm {0.6}}$ & 91.7$_{\pm {0.5}}$ & 58.8\phantom{$_{\pm {0.8}}$} & 95.9\phantom{$_{\pm {0.2}}$} & 78.7\phantom{$_{\pm {3.6}}$} & 91.8\phantom{$_{\pm {0.2}}$} \\
    
    DFR (Up) & \bxmark & 54.2$_{\pm {6.2}}$ & 90.3$_{\pm {2.0}}$ & \underline{89.9}$_{\pm {0.2}}$ & 91.3$_{\pm {0.3}}$ & 66.5\phantom{$_{\pm {0.8}}$} & 96.4\phantom{$_{\pm {0.9}}$} & 83.9\phantom{$_{\pm {2.3}}$} & 91.2\phantom{$_{\pm {0.8}}$} \\ 
    
    Orth-Cali & \bxmark & 74.0$_{\pm {0.0}}$ & 78.7$_{\pm {0.0}}$ & 82.2$_{\pm {0.0}}$ & 84.4$_{\pm {0.0}}$ & 68.8$_{\pm {0.0}}$ & 84.5$_{\pm {0.0}}$ & 76.1$_{\pm {0.0}}$ & 86.2$_{\pm {0.0}}$\\
    
    CA & \bxmark & \underline{82.5}$_{\pm {0.9}}$ & 88.2$_{\pm {2.6}}$ & 88.4$_{\pm {1.7}}$ & 90.8$_{\pm {1.2}}$ &85.3$_{\pm {2.3}}$ & 90.4\phantom{$_{\pm {0.2}}$} & 83.9\phantom{$_{\pm {1.1}}$} & 90.4\phantom{$_{\pm {0.0}}$}\\
    
    Yang~\etal~\cite{yang2023mitigating} & \bcmark & 77.5\phantom{$_{\pm {0.0}}$} & 83.2\phantom{$_{\pm {0.0}}$} & 75.2\phantom{$_{\pm {0.0}}$} & 80.4\phantom{$_{\pm {0.0}}$} & - & - & - & -\\

    CFR & \bxmark & 76.9\phantom{$_{\pm {0.0}}$} & 77.6\phantom{$_{\pm {0.0}}$} & 73.7\phantom{$_{\pm {0.0}}$} & 81.1\phantom{$_{\pm {0.0}}$} & - & - & - & - \\

    FairerCLIP & \bxmark & 75.4$_{\pm {1.9}}$ & 84.3$_{\pm {2.2}}$ & 81.5$_{\pm {0.7}}$ & 85.0$_{\pm {0.9}}$ & \underline{86.0}$_{\pm {1.8}}$ & 92.2$_{\pm {0.8}}$ & \underline{85.2}$_{\pm {2.3}}$ & 87.8$_{\pm {1.7}}$\\
    
    CoOPood & \bxmark & 60.3\phantom{$_{\pm {0.0}}$} & 82.4\phantom{$_{\pm {0.0}}$} & 31.1\phantom{$_{\pm {0.0}}$} & 78.1\phantom{$_{\pm {0.0}}$} & 75.2\phantom{$_{\pm {0.0}}$} & 90.7\phantom{$_{\pm {0.0}}$} & - & - \\
    
    \textbf{GroupCoOp}& \bxmark & \textbf{84.9}$_{\pm {1.1}}$ & 86.7$_{\pm {1.5}}$  
    & \textbf{90.3}$_{\pm {1.3}}$ & 92.2$_{\pm {0.3}}$ 
    & \textbf{87.6}$_{\pm {1.2}}$ & 90.7$_{\pm {1.9}}$
    & \textbf{89.4}$_{\pm {1.0}}$ & 90.5$_{\pm {0.6}}$\\
  \bottomrule
  \end{tabular}
  }
  \vspace{-2mm}
\end{table*}

\begin{table*}
  \caption{
  \textsc{Worst} and \textsc{InDist} metrics (\%) evaluated on Waterbirds using frozen CLIP RN101, CLIP ViT-B/16, and CLIP ViT-B/32. The best \textsc{Worst} is highlighted in \textbf{bold}.
  }
  \label{tab:performance_arch}
  \centering
  \begin{small}
  \begin{tabular}{lgcgcgcgc}
  \toprule
   & \multicolumn{2}{c}{CLIP RN101} & \multicolumn{2}{c}{CLIP ViT-B/16} & \multicolumn{2}{c}{CLIP ViT-B/32} \\
    Method & \textsc{Worst} & \textsc{InDist} & \textsc{Worst} & \textsc{InDist} & \textsc{Worst} & \textsc{InDist} \\
    \midrule
    Zero-shot (class)  & 33.6$_{\pm {0.0}}$ & 90.0$_{\pm {0.0}}$ & 34.0$_{\pm {0.0}}$ & 88.1$_{\pm {0.0}}$ & 47.0$_{\pm {0.0}}$ & 88.8$_{\pm {0.0}}$ \\
    CA & 82.0$_{\pm {1.3}}$ & 86.0\phantom{$_{\pm {0.0}}$} & 83.1$_{\pm {2.1}}$ & 90.9\phantom{$_{\pm {0.0}}$} & 80.7$_{\pm {1.4}}$ & 84.2\phantom{$_{\pm {0.0}}$} \\
    
    \textbf{GroupCoOp} & \textbf{84.7}$_{\pm {2.0}}$ & 89.3$_{\pm {0.5}}$ 
    & \textbf{86.2}$_{\pm {1.6}}$ & 89.2$_{\pm {0.7}}$ 
    & \textbf{81.3}$_{\pm {1.0}}$ & 85.5$_{\pm {2.9}}$\\
  \bottomrule
  \end{tabular}
  \end{small}
\end{table*}
\begin{table*}[th!]
  \begin{minipage}[t]{0.31\textwidth}
  \setlength\tabcolsep{2pt}
  \caption{Performance in \textsc{Worst}, and \textsc{InDist} (\%) on the Metashift dataset using frozen CLIP RN50. 
  }
  \vspace{-2mm}
    \begin{center}
    \begin{small}
    \resizebox{\linewidth}{!}{
        \begin{tabular}{lcgc}
        \toprule
        & Bias & \multicolumn{2}{c}{Metashift} \\
        Method&label&\textsc{Worst} & \textsc{InDist}\\
        \midrule
        ERM & \bxmark &73.9&90.1\\
        KNN & \bxmark &80.0 &90.1\\
        CoOp & \bxmark & \underline{84.1} & 93.2\\
        CoOp-CSC& \bxmark &83.1&92.9\\
        GroupDRO& \bcmark &83.2&87.3\\
        DFR& \bcmark &83.1&88.3\\
        AFR~& \bxmark &76.9&86.8\\
        JTT& \bxmark &78.5&89.4\\
        CnC& \bxmark &78.3&87.1\\
        CA~& \bxmark &77.9&85.5\\
        Yang~\etal~\cite{yang2023mitigating} & \bcmark &81.5&88.8\\
        CFR & \bxmark &81.5&89.5\\
        \textbf{GroupCoOp}& \bxmark &\textbf{89.2} &92.2\\
        \bottomrule
        \end{tabular}
    }
    \end{small}
    \end{center}
    \label{tab:performance_metashift}
  \end{minipage}
  \hfill
  \begin{minipage}[t]{0.65\textwidth}
  \centering
  \small
  \caption{\textsc{Worst} and \textsc{InDist} metrics (\%) evaluated on BREEDS using frozen CLIP RN50. We mark the best and the second-best \textsc{Worst} in \textbf{bold} and \underline{underline}, respectively.}
  \resizebox{\linewidth}{!}{
  \begin{tabular}{lcgcgc}
  \toprule
   & Bias & \multicolumn{2}{c}{Living17} & \multicolumn{2}{c}{Nonliving26}\\
    Method & label & \textsc{Worst} & \textsc{InDist} & \textsc{Worst} & \textsc{InDist}\\
    \midrule
    Zero-shot (class) & \bxmark & 6.0$_{\pm {0.0}}$ & 86.7$_{\pm {0.0}}$ & 6.0$_{\pm {0.0}}$ & 72.3$_{\pm {0.0}}$\\
    
    Zero-shot (group) & \bxmark & 30.0$_{\pm {0.0}}$ & 90.6$_{\pm {0.0}}$ & 56.0$_{\pm {0.0}}$ & 87.1$_{\pm {0.0}}$\\

    KNN & \bxmark & 66.0$_{\pm {0.0}}$ & 91.9$_{\pm {0.0}}$ & 56.0$_{\pm {0.0}}$ &89.3$_{\pm {0.0}}$ \\ 
    
    ERM Linear  & \bxmark & 53.3$_{\pm {0.9}}$ & 90.8$_{\pm {0.0}}$ & 32.0$_{\pm {0.0}}$ & 82.3$_{\pm {0.1}}$ \\ 
    
    ERM Adapter & \bxmark & 70.7$_{\pm {0.9}}$ & 94.0$_{\pm {0.1}}$ & 61.3$_{\pm {1.9}}$ & 92.1$_{\pm {0.2}}$\\ 
    WiSE-FT & \bxmark & 53.3$_{\pm {0.9}}$ & 90.8$_{\pm {0.0}}$ & 36.7$_{\pm {0.9}}$ & 83.6$_{\pm {0.1}}$\\
        
    CoOp & \bxmark & \underline{72.0}$_{\pm {6.0}}$ & 93.4$_{\pm {0.4}}$ & 65.3$_{\pm {3.1}}$ & 90.4$_{\pm {0.2}}$\\
    
    CoOp-CSC & \bxmark & 67.3$_{\pm {2.3}}$ & 93.5$_{\pm {0.1}}$ & \textbf{67.3}$_{\pm {1.2}}$ & 91.4$_{\pm {0.2}}$\\
    
    DFR (Sub) & \bxmark & 46.7$_{\pm {3.4}}$ & 89.3$_{\pm {0.3}}$ & 29.3$_{\pm {1.9}}$ & 80.6$_{\pm {0.1}}$\\
    
    DFR (Up) & \bxmark & 44.0$_{\pm {0.0}}$& 86.4$_{\pm {0.0}}$ & 30.0$_{\pm {4.1}}$ & 83.6$_{\pm {0.0}}$\\ 
    
    CA & \bxmark & 62.0$_{\pm {1.7}}$ & 90.9$_{\pm {0.3}}$ & 55.3$_{\pm {4.2}}$ & 88.1$_{\pm {0.6}}$\\

    \textbf{GroupCoOp}& \bxmark & \textbf{76.0}$_{\pm {2.0}}$ & 93.5$_{\pm {0.3}}$ & \textbf{67.3}$_{\pm {3.1}}$ & 91.1$_{\pm {0.2}}$\\
    
  \bottomrule
  \end{tabular}
  \label{tab:performance_breeds}
  }
  \end{minipage}
  \vspace{-4mm}
\end{table*}

\begin{table}[th!]
\centering
\small
\caption{Ablation studies using \textsc{Worst} accuracy (\%) using frozen CLIP RN50. We study the impact of group-wise classification (row 2) and prompt learning (row 3). We use linear probing when prompt learning is not applied. All experiments are conducted with upsampling.}
\setlength\tabcolsep{2.5pt}
\begin{tabular}{cccccc}
    \toprule
    \multicolumn{3}{c}{Method} & \multicolumn{3}{c}{Dataset}\\
    \cmidrule(lr){1-3}\cmidrule(lr){4-6}
    Linear & Group-wise & Prompt &\multirow{2}{*}{Waterbirds} & \multirow{2}{*}{CelebA} & BREEDS\\
    probing & classification & learning & & & Living17\\
    \midrule
    \bcmark & & & 81.3 & 88.1 & 69.3\\
    \bcmark & \bcmark & & 84.7 & 90.0 & 75.3\\
    & \bcmark  & \bcmark & 84.9 & 90.3 & 76.0\\
     \bottomrule    
\end{tabular}\label{tab:ablation}
\end{table}
\subsection{Quantitative results}
GroupCoOp achieved state-of-the-art results on the five real-world datasets across five CLIP variants.
In Table~\ref{tab:performance}, we observe that GroupCoOp outperforms existing debiased fine-tuning methods by a large margin on Waterbirds and CelebA using CLIP RN50 and CLIP ViT-L/14 as a frozen backbone. This demonstrates that simple group-wise prompt learning is more effective than retraining projection layers for improving group robustness. 
On the other heads, the poor performance of LWBC, DFR, and GroupDRO highlights the limitation of applying traditional debiasing algorithms to fine-tune frozen VLMs.
Additionally, in Table~\ref{tab:performance_arch}, we demonstrate the effectiveness of GroupCoOp three additional CLIP architectures on Waterbirds. GroupCoOp consistently outperformed CA across five CLIP architectures, suggesting its robustness across all five architectures. 
Table~\ref{tab:performance_metashift} shows results on the Metashift dataset. GroupCoOp is 9.8\% better than CFR, the previous state of the art. Compared with CoOp and CoOp-CSC, learning group-wise prompts is more effective.
In Table~\ref{tab:performance_breeds}, GroupCoOp achieved state-of-the-art results on BREEDS Living17 and Nonliving26, which suggests that GroupCoOp not only effectively mitigates biases present in the dataset but also adapts VLMs well to downstream tasks even when biases are not prominent in the training set.
Additionally, GroupCoOp achieved comparable performance even with full fine-tuning debiased training method on the multiple biases benchmark, MultiCelebA, demonstrating its effectiveness even when the number of samples from minority groups is extremely small and multiple subgroups exist for each class. The results are provided in Section~\ref{sec:multiceleba_results}.

\subsection{Ablation study}
\textbf{Effectiveness of each component.}
We conduct ablation studies to assess the impact of group-wise classification and prompt learning in our method. Specifically, we compare prompt learning with linear probing and examine the effect of training with group-wise classification versus the original classification. To ensure a controlled evaluation, upsampling is applied across all experiments in Table~\ref{tab:ablation}.
Comparing row 1 and row 2, we observe performance improvements across all three datasets when group-wise classification is applied. Furthermore, comparing row 2 and row 3, we find that prompt learning further enhances performance. These results validate the effectiveness of both components. A more in-depth analysis of each component is provided in the following sections.

\begin{table}[t]
  \caption{Comparisons between training a target class classifier and training a group classifier. All
experiments on group-wise learning are conducted using ground truth group labels.
  }
  \label{tab:baselines}
  \centering
  \setlength\tabcolsep{2.5pt}
  \begin{small}
  \begin{tabular}{llgcgc}
  \toprule
  &&\multicolumn{2}{c}{Waterbirds} & \multicolumn{2}{c}{CelebA}\\
     & & \textsc{Worst} & \textsc{InDist} & \textsc{Worst} & \textsc{InDist}\\
    \midrule
    \multirow{2}{*}{\textbf{Zero-shot}} 
    & Class-wise & 36.6 & 92.2 & 74.0 & 81.9\\
    & Group-wise & 55.9 & 87.8 & 70.8 & 82.6\\
    \midrule
    \multirow{2}{*}{\textbf{\shortstack[l]{Average\\feature}}}
    & Class-wise & 33.6 & 94.6 & \phantom{0}6.1 & 79.1\\
    & Group-wise & 70.4 & 87.2 & 71.0 & 81.8\\
    \midrule
    \multirow{2}{*}{\textbf{\shortstack[l]{ERM\\linear}}} 
    & Class-wise & 61.2 & 96.4 & 44.6 & 95.4\\ 
    & Group-wise & 69.6 & 95.2 & 62.8 & 94.8\\
    \midrule
    \multirow{2}{*}{\textbf{\shortstack[l]{ERM\\adapter}}}
    & Class-wise & 63.0 & 96.0 & 41.9 & 94.7\\ 
    & Group-wise & 81.0 & 96.4 & 78.9 & 93.2 \\ 
    \midrule
    \multirow{3}{*}{\textbf{\shortstack[l]{Prompt\\ learning}}}
    & CoOp &  63.4 & 96.3 & 44.3 & 95.2\\
    & CoOp-CSC & 65.0 & 96.1 & 43.5 & 95.2\\
    & GroupCoOp w/ GT & \textbf{86.1} & 87.7 & \textbf{90.4} & 92.4 \\
  \bottomrule
  \end{tabular}
  \end{small}
  \vspace{-2mm}
\end{table}

\textbf{Effectiveness of group classification.} 
We study the impact of group classification on group robustness. 
In Table~\ref{tab:baselines}, we compare class-wise training and group-wise training with four baselines: 
Zero-shot classification, ERM Linear Probe, ERM Adapter, and Prompt learning. 
For group-wise training, we used ground truth group labels and adopted the same inference procedure as GroupCoOp.
Group-wise training consistently exhibited better \textsc{Worst} across all cases, except for Zero-shot classification on CelebA, underscoring the effectiveness of learning group classification even with ERM.

\begin{figure}
\centering
\includegraphics[width=0.7\linewidth]{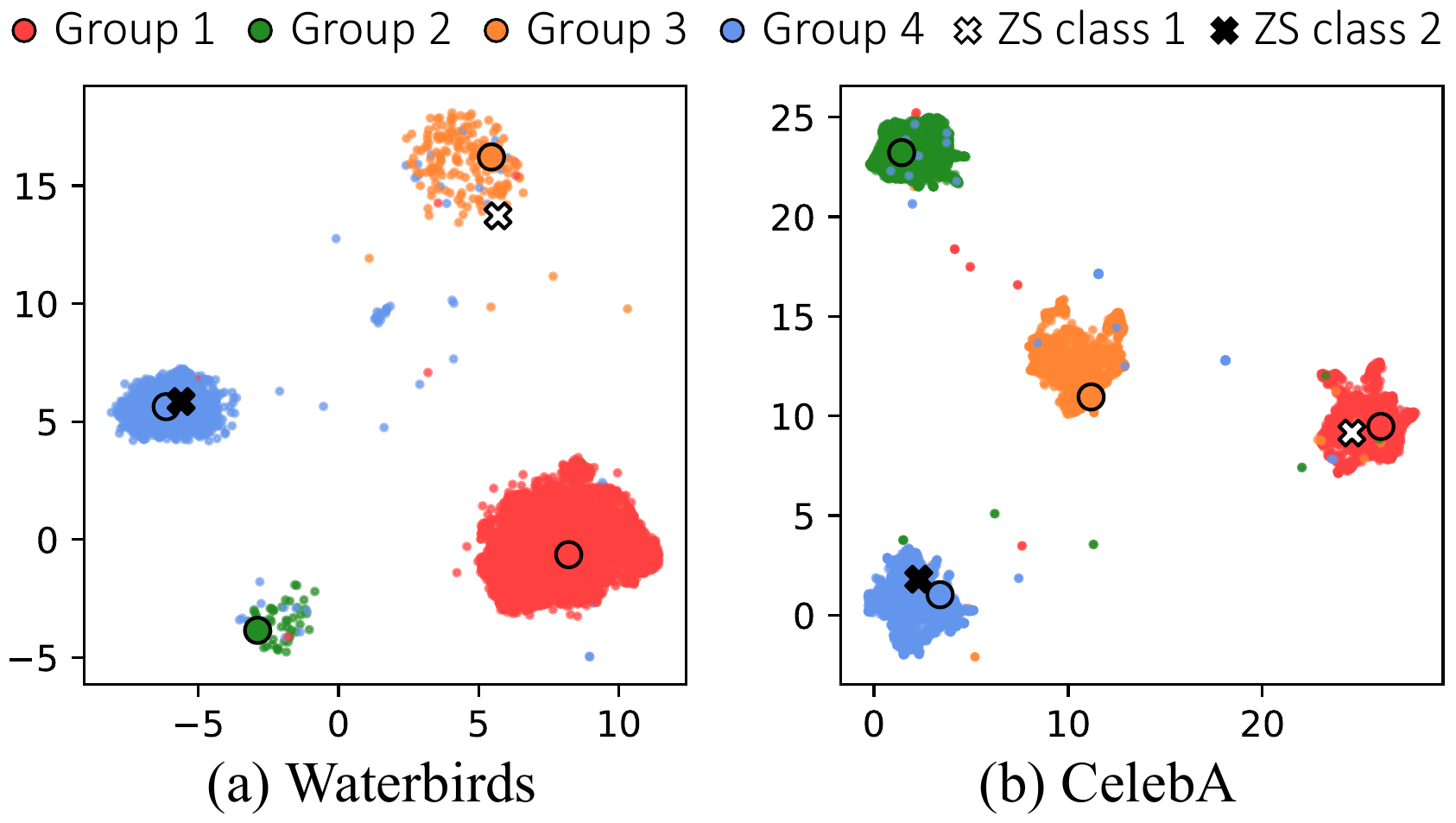}
\caption{
UMAP visualization of the text feature of learned group prompts (marked with outlined circles), predefined text prompts which are used for zero-shot inference (marked with crosses), and image features of the Waterbirds and CelebA training sets in the CLIP RN50 embedding space (marked with dots).
Each color represents a different group.
}
\label{fig:qual}
\end{figure}

\begin{table}[t!]
\vspace{-2mm}
  \caption{In-depth comparison between linear probing (LP) and prompt learning, focusing on the effect of the training samples size on \textsc{Worst} accuracy on Waterbirds using frozen CLIP RN50. All experiments were conducted on a subset of the original training set while maintaining the sample ratio of groups. Ground-truth group labels were used in all cases.
  }
  \label{tab:the_number_of_samples}
  \vspace{2mm}
  \centering
  \small
  \begin{tabular}{@{}lcccc@{}}
  \toprule
  & 5\% & 25\% & 50\% & 100\%\\
    \midrule
    GroupCoOp w/ LP & 64.2$_{\pm {5.3}}$ & 73.2$_{\pm {4.4}}$ 
    & 77.0$_{\pm {6.4}}$ & 85.4$_{\pm {3.0}}$\\
    GroupCoOp & 67.7$_{\pm {0.9}}$ & 77.3$_{\pm {1.8}}$ 
    & 80.4$_{\pm {3.1}}$ & 86.1$_{\pm {0.9}}$\\
  \bottomrule
  \end{tabular}
\end{table}
\textbf{Efficacy of prompt learning over linear probing.}
We hypothesize that prompt learning leverages language knowledge embedded in the text encoder, allowing it to maintain stable and superior performance even with an extremely small number of training samples.
To validate its effectiveness compared to linear probing (LP), we study how \textsc{Worst} accuracy varies with the number of training samples, using subsampling the original training set. With only 5\% training samples, the smallest group contained just three samples.
As shown in Table~\ref{tab:the_number_of_samples}, LP consistently underperforms and exhibits higher variance compared with prompt learning, with the performance gap widening as the sample size decreased. These results highlight the robustness of prompt learning in low-data regimes.
Beyond sample efficiency, LP can limit the generalizability of VLMs, as it requires retraining whenever new classes are introduced. In contrast, prompt learning enhances flexibility by enabling direct inference on unseen classes and efficient adaptation through new prompts. This ability to preserve the VLMs’ generalizability further highlights the advantages of prompt learning over LP.

\begin{table}[t!]
\vspace{-2mm}
\caption{
Comparison of the number of trainable parameters on waterbirds using CLIP RN50. The relative number of parameters compared to the Resnet50 is reported in the third column.
}
\begin{center}
\begin{small}
\begin{tabular}{lrr}
    \toprule
    Method & \# Trained params & \% Params \\
    \midrule
    Resnet50 & 25,557,032 & 100.000\\
    \midrule
    CA & 263,424 & 1.031 \\
    Yang~\etal & 14,789,632 & 57.896\\
    CFR & 14,789,632 & 57.896\\
    CoOPood & 1,026,000 & 4.015 \\ 
    GroupCoOp & 4,096 & 0.016 \\
    \bottomrule
\end{tabular}
\end{small}
\end{center}
\label{tab:computational_cost}
\vspace{-2mm}
\end{table}
\textbf{Comparison of trainable parameters.}
To evaluate the efficiency of our method, we compare the number of trained parameters using the CLIP RN50 backbone and its relative proportion to the full fine-tuning of Resnet50.
As shown in Table~\ref{tab:computational_cost}, GroupCoOp requires only 4,096 trainable parameters, which is a mere 0.016\% of the parameters of ResNet50. 
In comparison, Yang~\etal and CFR require significantly more parameters, as they involve training projection layers of the CLIP image encoder. 
Although CA and CoOPood use relatively fewer parameters than these methods, GroupCoOp utilizes significantly fewer parameters and outperforms them by a large margin in terms of performance. These results underscore the efficiency of GroupCoOp, providing a lightweight solution while maintaining outstanding performance.

\subsection{Analysis of learned group prompts}
To empirically verify whether the learned group prompt effectively represents each group, we visualized the text features of the learned group prompts marked with outlined circles, the text features of predefined text prompts which are used for zero-shot inference (denoted as `ZS class \#'), and the image feature of the Waterbirds and CelebA training set using UMAP in the CLIP-RN50 embedding space. 
As shown in Figure~\ref{fig:qual}, 
predefined text prompts represent only specific groups, which can introduce biases and lead to less accurate representations. 
In contrast, the text features derived from each learned group prompt are clustered with the image features of their corresponding group in the embedding space, confirming that each prompt effectively represents its group. 

\section{Conclusion}
We introduced GroupCoOp, a novel debiased fine-tuning algorithm for VLMs that enhances group robustness without explicit group annotations. GroupCoOp learns group-specific text prompts and uses a pseudo group labeling strategy to address group imbalances. Our evaluations across five benchmarks show that GroupCoOp outperforms existing methods with only 0.016\% of parameters updated, demonstrating efficiency and effectiveness. Notably, GroupCoOp excels even with limited training samples, highlighting its robustness. 
This work contributes to fairer and more robust VLM adaptations in a variety of downstream tasks.

\bibliography{egbib}
\bibliographystyle{plain}
\appendix
\newpage 
\section{Quantitative results in multiple spurious correlations setting}\label{sec:multiceleba_results}
\begin{table*}[h!]
\vspace{-2mm}
\caption{
Performance in \textsc{Worst}, and \textsc{InDist} (\%) on MultiCelebA in two biases setting (\texttt{gender} and \texttt{age}).
We mark the best and the second-best performance in \textbf{bold} and \underline{underline}, respectively.
}
\begin{center}
\begin{small}
\begin{tabular}{clcgc}
\toprule
\multirow{2}{*}{Backbone} & \multirow{2}{*}{Method} & Group label of&\multicolumn{2}{c}{MultiCelebA}\\ 
&&train set used & \textsc{Worst} & \textsc{InDist}\\
\midrule
\multirow{8}{*}{\shortstack{ImageNet\\pretrained\\ RN50}} & ERM & \bxmark  & 14.7$_{\pm {4.8}}$  & 97.0$_{\pm {0.2}}$\\
& Upsampling & \bcmark  & 71.5$_{\pm {2.0}}$ & 82.6$_{\pm {0.8}}$ \\
& Upweighting & \bcmark & \underline{73.5}$_{\pm {4.2}}$ & 83.4$_{\pm {5.9}}$ \\
& GroupDRO &\bcmark & 71.6$_{\pm {1.1}}$ & 83.5$_{\pm {0.7}}$\\
& SUBG& \bcmark & 69.6$_{\pm {0.7}}$& 80.3$_{\pm {1.1}}$\\
& LISA & \bcmark & 72.8$_{\pm {1.5}}$ & 84.5$_{\pm {1.7}}$\\
& DFR$_{tr}^{tr}$ & \bcmark & 12.3$_{\pm {8.5}}$ & 85.5$_{\pm {6.2}}$\\ 
& Kim~\etal~\cite{kim2023removing} &\bcmark & \textbf{77.9}$_{\pm {0.2}}$ & 84.3$_{\pm {0.9}}$\\ 
\midrule
\multirow{4}{*}{\shortstack{Frozen\\CLIP\\RN50}} & Zero-shot (class) &\bxmark & 3.1$_{\pm {0.0}}$ & 31.1$_{\pm {0.0}}$\\
& ERM Linear (class) & \bxmark & 23.1$_{\pm {1.2}}$ & 95.5$_{\pm {0.6}}$\\
& ERM Linear (group) & \bcmark & 17.4$_{\pm {0.8}}$ & 96.1$_{\pm {0.2}}$\\
& \textbf{GroupCoOp} with GT &\bcmark & \textbf{77.8}$_{\pm {0.6}}$ & 81.8$_{\pm {0.1}}$\\
\bottomrule
\end{tabular}
\end{small}
\end{center}
\label{tab:multiceleba_table}
\end{table*}
\noindent\textbf{MultiCelebA}~\citep{CelebA,kim2023removing}.
MultiCelebA benchmarks multiple bias types; among them, we adopt the two-bias setting (age and gender) of MultiCelebA, in which each class consists of 4 subgroups ($\mathcal{A}=\{\{$Young \& Female$\}$, $\{$Young \& Male$\}$, $\{$Old \& Female$\}$, $\{$Old \& Male$\}\}$). Compared with single-bias benchmarks, the number of samples for minority groups is much lower, and the groups exhibit multiple spurious correlations, making training more challenging. 

\noindent\textbf{Results. }
Table~\ref{tab:multiceleba_table} presents the \textsc{Worst} and \textsc{InDist} metrics, evaluated on the MultiCelebA dataset under two bias settings (gender and age), for various methods including GroupCoOp. 
The methods using the Resnet50 backbone involve full fine-tuning of ImageNet pretrained models, while the methods using the CLIP RN50 backbone involve freezing the CLIP model and learning only a small number of additional parameters.
GroupCoOp achieves competitive performance in both \textsc{WORST} and \textsc{InDist}. 
GroupCoOp achieves competitive performance in both \textsc{WORST} and \textsc{InDist}. Specifically, GroupCoOp achieved a \textsc{WORST} score of 77.8\% even with frozen backbone, which is comparable to the highest score achieved by Kim~\etal~\cite{kim2023removing} at 77.9\%, which utilized ground truth bias labels.
Notably, GroupCoOp achieves this performance with only 0.016\% of the learnable parameters compared to Kim~\etal~\cite{kim2023removing}, highlighting the efficiency and effectiveness of our method.

\section{Comparison with full fine-tuning models.} 
\begin{table*}[t!]
\caption{
Comparison with full fine-tuning models. All experiments are conducted using Resnet50. The full fine-tuning methods trained all parameter of ImageNet pretrained Resnet50, while parameter-efficient fine-tuning methods trained a small number of parameters atop the frozen CLIP Resent50.
}
\begin{center}
\begin{small}
\begin{tabular}{lccgcgc}
\toprule
\multirow{2}{*}{Method}&Trained &Group label of &\multicolumn{2}{c}{Waterbirds} & \multicolumn{2}{c}{CelebA}\\
& params (\%) & train set used & \textsc{Worst} & \textsc{InDist} & \textsc{Worst} & \textsc{InDist} \\
\midrule
GroupDRO &100 & \bcmark & 89.9 & 92.0 & 88.9 & 93.9\\ 
SSA & 100 &\bcmark & 89.0 & 92.2 & \underline{89.8} & 92.8 \\
DFR$_{tr}^{tr}$ &  100 & \bcmark  & 90.2 & 97.0 & 80.7 & 90.6\\ 
DFR$_{tr}^{val}$ &  100 & \bcmark  & \underline{92.9} & 94.2 & 88.3 & 91.3\\ 
\midrule
Yang~\etal~\cite{yang2023mitigating} & 57.90 & \bcmark & 77.5 & 83.2 & - & - \\
CoOPood with GT & 4.015 & \bcmark & 85.6 & 89.7 & 72.2 & 88.8 \\
\textbf{GroupCoOp with GT}& \textbf{0.016} & \bcmark & \textbf{86.1}$_{\pm {0.9}}$ & 87.7$_{\pm {0.3}}$ & \textbf{90.4}$_{\pm {1.8}}$ & 92.4$_{\pm {0.2}}$\\ 

\midrule
ERM & 100  & \bxmark & 63.7 & 97.0 & 47.8 & 94.9 \\
LfF & 100  & \bxmark & 78.0 & 91.2 & 70.6 & 86.0\\ 
EIIL & 100  & \bxmark & 77.2& 96.5 & 81.7 & 85.7\\
JTT & 100  & \bxmark  & 83.8 & 89.3 & 81.5 & 88.1\\
CNC & 100 & \bxmark & 88.5 & 90.9  & 88.8 & 89.9\\ 
DISC & 100 & \bxmark & \underline{88.7} & 93.8& - & -\\
\midrule
ERM Adapter & 1.031 & \bxmark & 60.8 & 96.0 & 36.1 & 94.2 \\
CA & 1.031 & \bxmark & 83.7 & 89.4 & \underline{90.0} & 90.7 \\
CoOPood & 4.015 & \bxmark & 60.3 & 82.4 & 31.1 & 78.1\\
\textbf{GroupCoOp} & \textbf{0.016} & \bxmark & \textbf{84.9}$_{\pm {1.1}}$ & 86.7$_{\pm {1.5}}$  
& \textbf{90.3}$_{\pm {1.3}}$ & 92.2$_{\pm {0.3}}$
\\

\bottomrule
\end{tabular}
\end{small}
\end{center}
\label{tab:full_fine_tuning}
\end{table*}

We evaluated GroupCoOp against prior debiased training algorithms, which require training the entire parameter set of an ImageNet-pretrained ResNet50 on Waterbirds and CelebA. Note that both encoders of CLIP-RN50 were never trained on either dataset.
For a comparison with debiased training algorithms that require group labels, GroupCoOp utilized ground truth group labels, denoted as \textit{GroupCoOp with GT}.
As shown in Table~\ref{tab:full_fine_tuning}, GroupCoOp outperformed debiased training algorithms on the CelebA dataset and showed decent performance on the waterbirds dataset even utilizing only 0.016\% of their parameters. 
This superior results is attributed to GroupCoOp's effective utilization of language knowledge, enabling it to achieve higher accuracy with fewer parameters.

\section{Effectiveness of pseudo group labeling.}\label{sec:effectiveness_of_pseudo}
\begin{table*}[t]
  \caption{Ablation studies on pseudo group labeling strategy. GroupCoOp with pseudo group labels using (a) Ground truth, (b) zero-shot prediction, (c) K-means clustering, and (d) Ours.
  }
  \label{tab:ablation_pseudo}
  \centering
  \small
  \begin{tabular}{lgcgcgc}
  \toprule
  &\multicolumn{2}{c}{Waterbirds} & \multicolumn{2}{c}{CelebA} & \multicolumn{2}{c}{BREEDS Living17} \\
    Pseudo labeling & \textsc{Worst} & \textsc{InDist} & \textsc{Worst} & \textsc{InDist} & \textsc{Worst} & \textsc{InDist} \\
    \midrule
    (a) Ground truth & \textbf{86.1}$_{\pm {0.9}}$ & 87.7$_{\pm {0.3}}$ & \textbf{90.4}$_{\pm {1.8}}$ & 92.4$_{\pm {0.2}}$ & \textbf{76.7}$_{\pm {2.3}}$ & 94.0$_{\pm {0.2}}$ \\
    (b) ZS prediction & 63.9$_{\pm {4.3}}$ & 94.7$_{\pm {0.4}}$ & 56.7$_{\pm {4.5}}$ & 88.1$_{\pm {1.5}}$ & 68.7$_{\pm {4.2}}$ & 91.8$_{\pm {0.0}}$\\
    (c) K-means clustering & 65.1$_{\pm {0.6}}$ & 94.8$_{\pm {0.5}}$ & 73.1$_{\pm {2.2}}$ & 92.9$_{\pm {0.2}}$ & 69.3$_{\pm {3.1}}$ & 92.9$_{\pm {0.2}}$\\
    (d) Ours & \underline{84.9}$_{\pm {1.1}}$ & 86.7$_{\pm {1.5}}$  
    & \underline{90.3}$_{\pm {1.3}}$ & 92.2$_{\pm {0.3}}$ 
    & \underline{76.0}$_{\pm {2.0}}$ & 93.5$_{\pm {0.2}}$ \\
  \bottomrule
  \end{tabular}
\end{table*}

\begin{figure*}[h]
\centering
\includegraphics[width=0.7\linewidth]{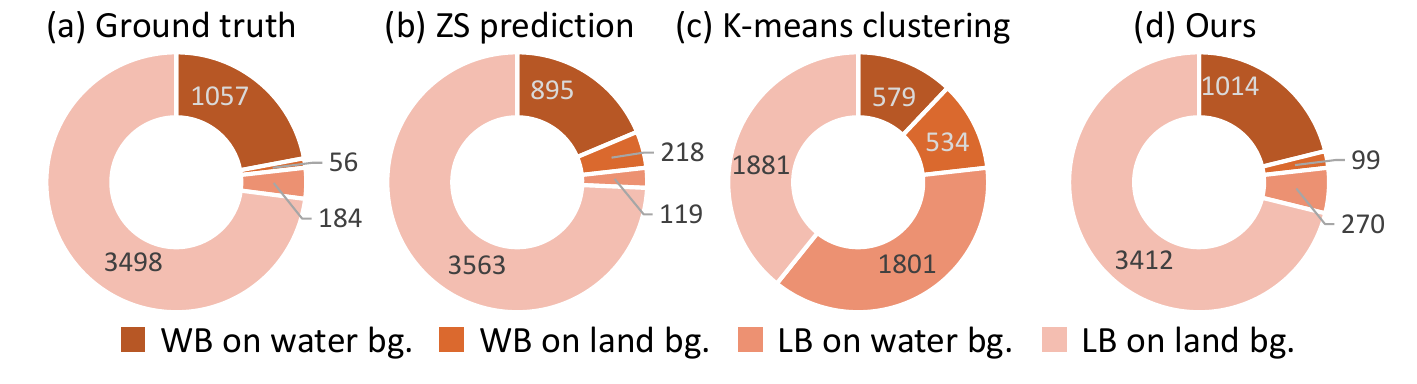}
\caption{The population of groups with ground truth group labels and three types of pseudo group labels in Waterbirds training set.}
\label{fig:stats_pseudo_label}
\end{figure*}

To verify the impact of pseudo group labeling, Figure~\ref{fig:stats_pseudo_label} presents the distribution of training samples across groups, and Table~\ref{tab:ablation_pseudo} compares our pseudo group labeling with three alternatives: 
(a) Zero-shot prediction for pseudo group labeling, as used in CA~\citep{zhang2022contrastive}, grouping samples based on the correctness of zero-shot classification for each class. 
(b) K-means clustering (K=2) for each class. 
(c) Ground truth group labels. 
(d) Average feature of each class in the visual-language embedding space.
Interestingly, (a) exhibited worse performance compared to (d). 
We hypothesize that this discrepancy arises from the zero-shot classifier being constructed solely using class names and predefined prompts, failing to capture the data distribution crucial for downstream tasks.
Moreover, (b) showed inferior performance on Waterbirds, as K-means clustering failed to detect group imbalance within each class, as shown in Figure~\ref{fig:stats_pseudo_label}. 
In contrast, (d) demonstrated comparable or even superior performance compared to ground truth group labels and effectively captured group imbalance within each class.

\end{document}